\documentclass[lettersize,journal]{IEEEtran}
\usepackage{amsmath,amsfonts}
\usepackage{algorithmic}
\usepackage{algorithm}
\usepackage{booktabs} 
\usepackage{array}
\usepackage[caption=false,font=normalsize,labelfont=rm,textfont=rm]{subfig}
\usepackage{textcomp}
\usepackage{float}
\usepackage[ colorlinks = true,
linkcolor = blue,
urlcolor  = blue,
citecolor = red,
anchorcolor = green,]{hyperref}

\usepackage{verbatim}
\usepackage{graphicx}
\usepackage{epstopdf}
\usepackage{cite}
\usepackage{adjustbox}
\usepackage{float}
\usepackage{multirow}
\usepackage{pifont}
\usepackage{amssymb}
\usepackage{dblfloatfix}  
\usepackage{graphicx}
\usepackage{dblfloatfix}  

\usepackage[table,xcdraw]{xcolor}
\begin{document}
\setlength{\textfloatsep}{5pt}

\title{UniDet-D: A Unified Dynamic Spectral Attention Model for Object Detection under Adverse Weathers}

\author{
    Wei Zhang*, Yuantao Wang*, Haowei Yang, Yin Zhuang, Shijian Lu, and Xuerui Mao\textsuperscript{\dag}
    \thanks{* Wei Zhang and Yuantao Wang contributed equally to this work.}
    \thanks{$\dagger$ Corresponding author: Xuerui Mao}
    \thanks{Wei Zhang is with the School of Interdisciplinary Science, Beijing Institute of Technology, Beijing 100081, China. (e-mail: w.w.zhanger@gmail.com).} 
    \thanks{Yuantao Wang and Haowei Yang are with the North China University of Technology, Beijing 100081, China. (e-mail: wyt.jackson@gmail.com, hackeryhw00@gmail.com)}
    \thanks{Yin Zhuang is with the National Key Laboratory of Science and Technology on Space-Born Intelligent Information Processing, Beijing Institute of Technology, Beijing 100081, China. (e-mail: yzhuang@bit.edu.cn).}
    \thanks{Shijian Lu is with the School of Computer Science and Engineering, Nanyang Technological University, Singapore. (e-mail: Shijian.Lu@ntu.edu.sg).}
    \thanks{Xuerui Mao is with the School of Interdisciplinary Science, Beijing Institute of Technology, Beijing 100081, China, and also with Beijing Institute of Technology (Zhuhai), Zhuhai, 519088, China (e-mail: maoxuerui@sina.com).} 
}

\maketitle

\begin{abstract}
Real-world object detection is a challenging task where the captured images/videos often suffer from complex degradations due to various adverse weather conditions such as rain, fog, snow, low-light, etc. Despite extensive prior efforts, most existing methods are designed for one specific type of adverse weather with constraints of poor generalization, under-utilization of visual features while handling various image degradations. Leveraging a theoretical analysis on how critical visual details are lost in adverse-weather images, we design UniDet-D, a unified framework that tackles the challenge of object detection under various adverse weather conditions, and achieves object detection and image restoration within a single network. Specifically, the proposed UniDet-D incorporates a dynamic spectral attention mechanism that adaptively emphasizes informative spectral components while suppressing irrelevant ones, enabling more robust and discriminative feature representation across various degradation types. 
Extensive experiments show that UniDet-D achieves superior detection accuracy across different types of adverse-weather degradation. Furthermore, UniDet-D demonstrates superior generalization towards unseen adverse weather conditions such as sandstorms and rain-fog mixtures, highlighting its great potential for real-world deployment.
\end{abstract}

\begin{figure}
	\centering
	\includegraphics[scale=0.24]{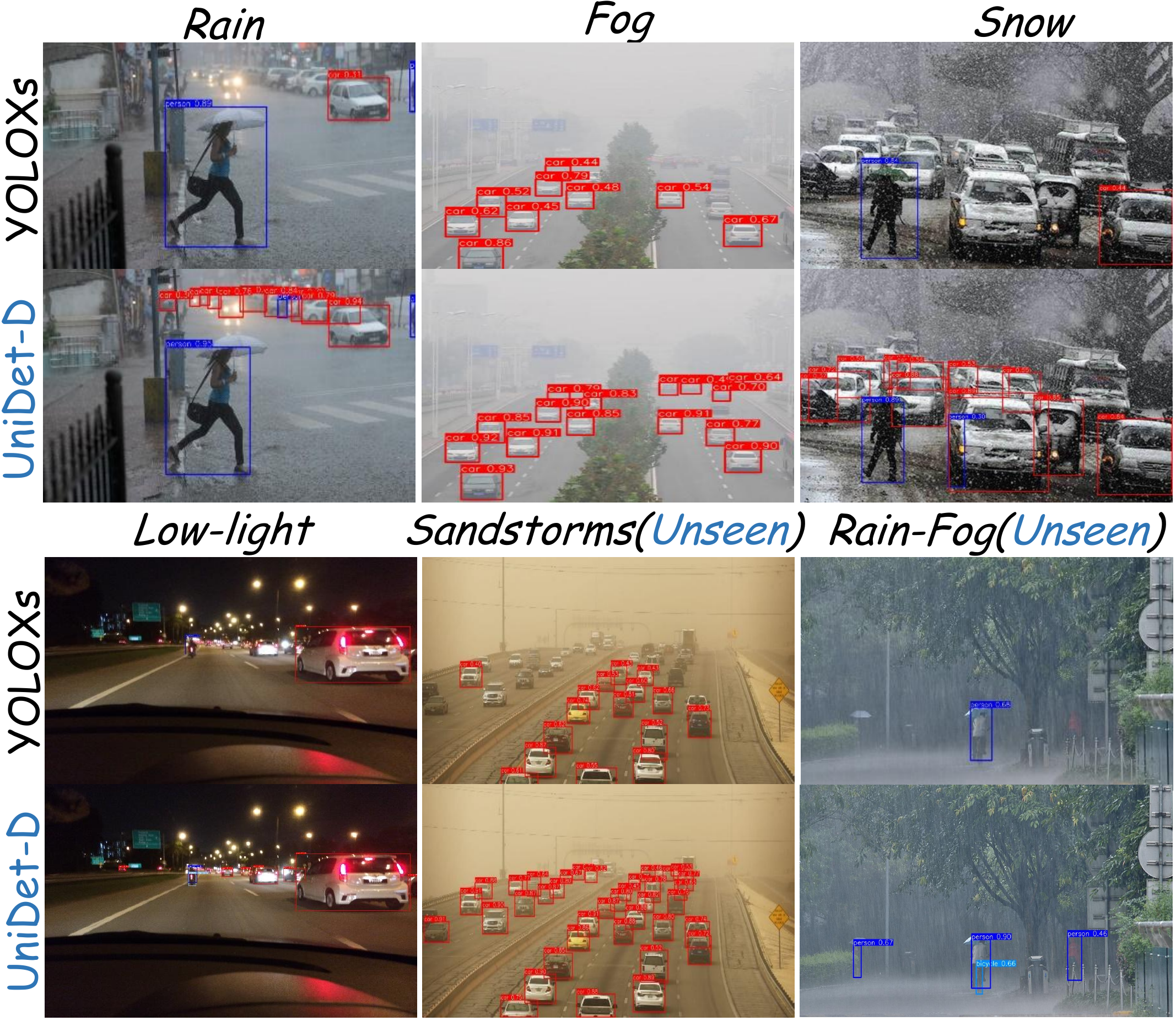}
	\caption{The proposed UniDet-D is a unified end-to-end model capable of handling diverse adverse weather conditions within a single framework. It outperforms the baseline detector YOLOXs~\cite{ge2021yolox} consistently across various adverse weather conditions, including rain, snow, fog, low light, sandstorms, and rain-fog conditions (the last two are unseen during training). \textit{\textbf{Note}: Bounding boxes are color-coded by object category: red for vehicles, dark blue for pedestrians, and light blue for bicycles.}}  
	\label{FIG:fig1}
\end{figure}

\section{Introduction}
\IEEEPARstart{O}{bject} {detection is a crucial aspect of environmental perception, aiming to accurately identify and locate objects of interest in images or videos~\cite{zhang2023perception,lee2022learning}. It serves as a foundation for key tasks such as path planning, motion prediction, decision-making, and control in autonomous driving scenarios~\cite{zou2023object,zhang2024earthmarker,zhang2024earthgpt}. Although deep learning has significantly advanced object detection~\cite{gao2022adaMixer}, adverse weather conditions such as rain, fog, snow, and low-light can severely degrade image quality, posing persistent challenges to the generalization and accuracy of existing mainstream models. This performance degradation ultimately damages the stability and functional robustness of intelligent visual systems in real-world deployments. Hence, researching effective low-quality imagery object detection under adverse weather conditions is a meaningful area of study.

Various studies have been conducted to enhance the reliability of degraded imagery object detection. These models can be broadly categorized into three main approaches: image adaptation, domain adaptation, and multi-task learning. The image adaptation focuses on enhance detection performance by adaptively improving image quality under degradation. Early works~\cite{liu2022IA-YOLO, liu2023Improving, Kalwar2023GDIP} use fixed-parameter filters, while later methods~\cite{xi2024detection, zhang2024cpa} introduce dynamic, context-aware enhancement to handle complex degradation patterns.
Domain adaptation approaches~\cite{oza2024Unsupervised, Li2022Cross} have been applied to object detection in adverse weather, aiming to learn domain-invariant features between high-quality (source domain) and low-quality (target domain) images. Additionally, the multi-task learning methods~\cite{wang2022togethernet, huang2021dsnet} employ multiple sub-networks to handle tasks such as image restoration and object detection, with some feature extraction layers shared between those sub-networks. 

Although the aforementioned methods have achieved certain progress, several significant challenges remain unaddressed. In particular, the existing approaches face limitations are summarized below: \textbf{(i)} From a framework design perspective, these methods are specifically tailored for individual weather degradations~\cite{zhong2024dehazing,wang2024mdd}, lacking a unified end-to-end solution capable of simultaneously handling diverse scenarios such as rain, snow, fog, and other complex degradations. \textbf{(ii)} In terms of representation learning, existing models fail to fully utilize informative signals within degraded images. In particular, they tend to exhibit a bias toward low-frequency or fixed-frequency signals~\cite{qin2021fcanet,xie2023boosting}. This frequency bias often leads to the underutilization of informative high-frequency details. Consequently, these limitations lead to suboptimal detection performance of object detectors under different degradations.

Motivated by the above observations and to address these limitations, we propose a \textbf{uni}fied object \textbf{det}ection model for various \textbf{d}egraded images, termed UniDet-D, which features a degradation-aware design based on dynamic spectral attention. As shown in real-world scenes in Fig. \ref{FIG:fig1}, the proposed UniDet-D unifies object detection under various adverse weather conditions (i.e., rain, snow, fog, low-light, sandstorms, and rain-fog). In addition, the proposed UniDet-D simultaneously integrates object detection and image restoration into a single end-to-end model. It eliminates the need to train separate models for different degraded scenarios. This unified design significantly improves generalization and efficiency, making UniDet-D a practical and scalable solution for real-world low-quality visual understanding tasks.

Concerning feature representation, we conduct an analysis of degraded images from a frequency-domain perspective to capture critical visual details lost in adverse conditions. Given that images are composed of diverse frequency components and that low- and high-frequency components both contain useful signals, and convolution neural networks (CNNs) tend to be more sensitive to low-frequency signals than high-frequency ones. To compensate for this inadequate utilization of high-frequency information in feature learning, we design a dynamic spectral perceive-select strategy. Specifically, a multi-spectrum perception mechanism (MSP) is proposed, which adaptively captures informative frequency components to enhance feature representation under degraded conditions. 
The proposed MSP module extracts frequency components of images through discrete cosine transform (DCT) processing~\cite{ahmed1974discrete}. Most importantly, unlike previous methods that rely on manually designed frequency selection~\cite{qin2021fcanet}, this paper introduces an adaptable frequency filtering ($\mathrm{AF^{2}}$) method, enabling the model to automatically adapt to informative spectral components. The effectiveness of our approach has been validated through extensive experiments compared with state-of-the-art (SOTA) detectors, where it achieves superior detection accuracy cross diverse degradation types and challenging weather conditions.

In summary, the contributions of this paper are as follows.

\begin{itemize}
\item{To the best of our knowledge, we are the first to propose a unified object interpretation model for various degraded scenario, termed UniDet-D, which is capable of jointly performing object detection and image restoration under a wide range of adverse weather conditions, such as rain, snow, fog, low-light, and sandstorms. UniDet-D contributes a versatile paradigm for object detection and restoration across different types of adverse weather, and therefore represents a significant advancement in the field of degraded visual understanding.}

\item{Building upon the theoretical insights of frequency utilization, a dynamic spectral perceive-select strategy is proposed, including an MSP module for thoroughly analyzing and capturing frequency components in degraded images, and a $\mathrm{AF^{2}}$ module adaptively emphasizes crucial frequencies while suppressing irrelevant ones. This design enhances the discriminative capacity of feature representations and enables adaptive object detection under unknown degradations.}

\item{Extensive experiments validate the efficacy of UniDet-D, compared to recent SOTA detectors in a variety of degradation scenarios. Moreover, UniDet-D demonstrates strong detection capability on even untrained adverse weather conditions such as sandstorms and rain-fog mixtures. These results highlight the generalization and practical potential of the model in real-world applications.}
\end{itemize}
}

\section{Related Works}
\label{sec2}
\subsection{General Object Detection}
Object detection aims to accurately locate and classify objects in images or videos. Current methods can fall into two major categories: region proposal-based object detection methods and regression-based object detection methods. Region proposal-based approaches, such as Libra R-CNN~\cite{pang2019libra}, and Dynamic R-CNN~\cite{zhang2020dynamic}, utilize selective search techniques~\cite{uijlings2013selective} to generate candidate regions that potentially contain objects in the first stage. In the second stage, they refine object localization and classification with these regions. Although these methods achieve high detection accuracy, their multi-stage process limits real-time performance. On the other hand, regression-based methods, such as SSD~\cite{liu2016ssd}, RetinaNet~\cite{Lin2017focal}, CenterNet~\cite{zhou2019objects}, and the YOLO series~\cite{ge2021yolox, wang2023yolov7, wang2025yolov9, wang2024yolov10}, eliminate the need to generate candidate bounding boxes. Instead, they directly detect and classify objects across the entire image, significantly improving computational efficiency. Although the above methods improve detection performance under good weather conditions, they may fall short in accurately locating and classifying objects under adverse weather conditions.

\subsection{Object Detection in Degraded Images}

Compared to clear-weather scenarios, object detection under adverse conditions has attracted significantly less research attention. Current approaches for object detection in degraded imagery fall into three main categories, as described below.

\begin{figure*}
	\centering		\includegraphics[scale=0.260]{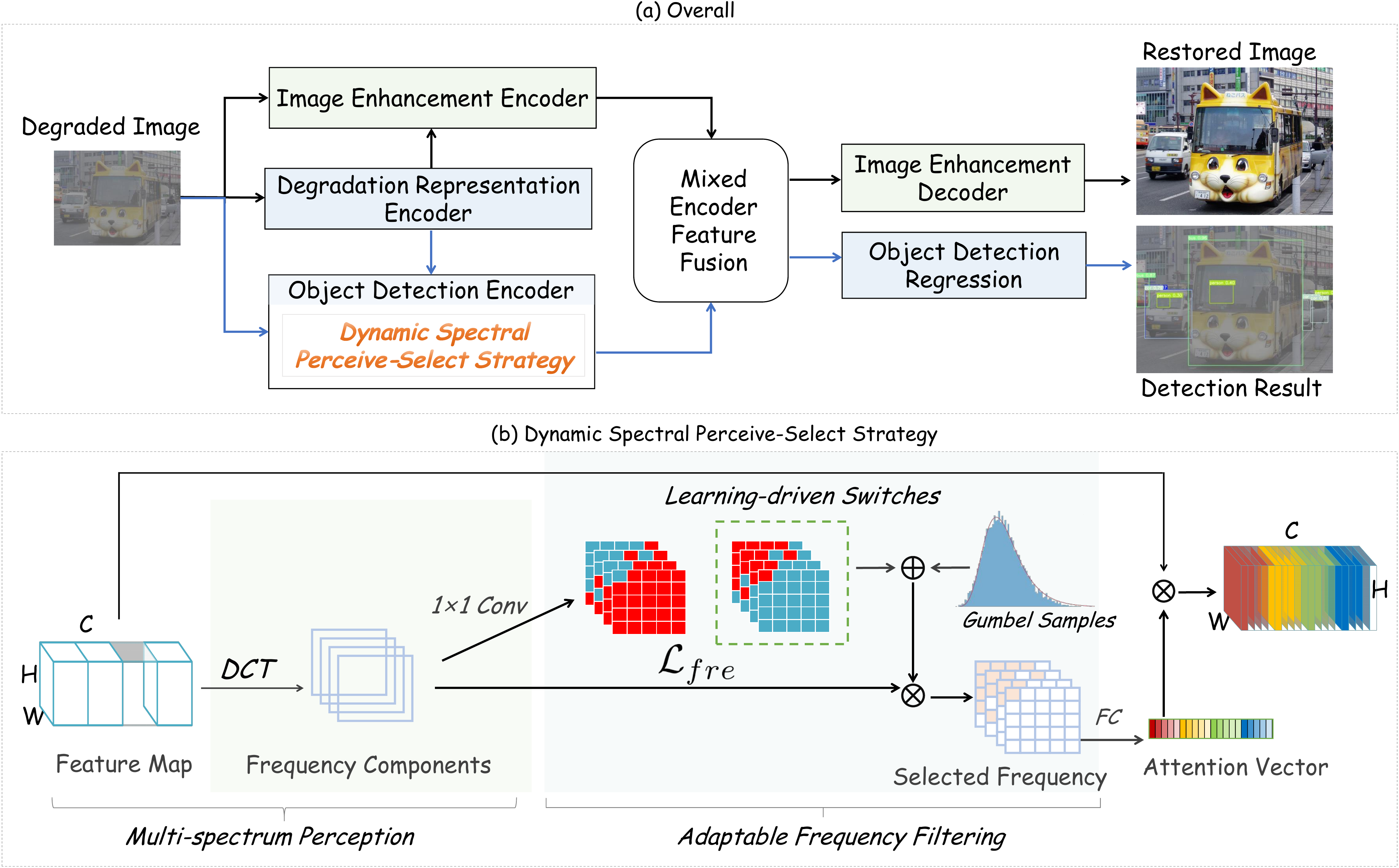}
	\caption{(a) Overall architecture of the proposed UniDet-D framework, which integrates image restoration and object detection in a unified manner under various degradation scenarios; (b) Illustration of the proposed dynamic spectral perceive-select strategy, which leverages multi-spectrum analysis and adaptable frequency filtering to enhance feature representation under diverse weather conditions.}
	\label{FIG:fig2}
\end{figure*}

\subsubsection{Image Adaptation}
Some works have explored adaptive enhancement to improve detection in degraded images by cascading enhancement and detection modules. For instance, IA-YOLO~\cite{liu2022IA-YOLO, liu2023Improving} introduces a differentiable image processing module that adjusts image quality under adverse conditions. MDD-ShipNet~\cite{wang2024mdd} extends this strategy to maritime scenarios, while GDIP~\cite{Kalwar2023GDIP} employs gated image filters for improved robustness. However, these methods often use fixed-parameter filters, limiting adaptability to spatially heterogeneous degradations. DEDet~\cite{xi2024detection} has then been proposed to overcome this, estimating pixel-wise filter parameters to dynamically adapt enhancement. Zhang et al.~\cite{zhang2024cpa} further proposed a CoT-guided adaptive enhancer that refines enhancement based on inferred degradation types. 

\subsubsection{Domain Adaptation} 
Recent efforts~\cite{zhang2022spectral, li2023domain, kennerley2023pcnet, wang2023ryolo, pang2024mcnet, huang2021fsdr} have explored domain adaptation to bridge the gap between clear and degraded image domains. For example, Robust-YOLO~\cite{wang2023ryolo} introduces an image transfer module to construct an intermediate domain for guiding adversarial training. MCNet~\cite{pang2024mcnet} employs a magnitude corrector to filter out irrelevant frequency content.
While effective to some extent, these methods often rely on training data diversity and overlook the degradation-induced distortion in feature extraction.

\subsubsection{Multi-task Approaches}
Another promising direction is joint learning of image restoration and object detection, which has shown improved performance on degraded images~\cite{cui2021multitask, xi2023coderainnet, huang2021dsnet, wang2022togethernet, xiao20233d}. DS-Net~\cite{huang2021dsnet} connects restoration and detection via shared convolution layers, allowing the detector to benefit from restored features. MAET~\cite{cui2021multitask} leverages degradation transformations to enhance structural understanding in detection. 
TogetherNet~\cite{wang2022togethernet} integrates both tasks using a dynamic transformer-based feature enhancer. RDMNet~\cite{wang2024RDMNet} further introduces degradation modeling to guide both subtasks, boosting adaptability across different weather types. Despite these advances, most methods still lack principled treatment of frequency information, which is vital for learning discriminative features under complex degradations.

\textit{Motivation and Open Challenges}.
Although progress has been made, two key issues remain. First, most methods over-rely on low-frequency signals, limiting the capture of fine-grained object details. Second, existing frameworks are often designed for single weather types, lacking scalability to diverse degradations. Therefore, a unified end-to-end solution is indispensable to jointly restore and detect objects under diverse adverse weather conditions while adaptively leveraging frequency-domain features for robust generalization. To this end, this paper introduces UniDet-D, a unified multi-task framework designed to effectively address these challenges.

\section{Methodology}
\label{sec3}
\subsection{Overview}
\label{subsec1}

The overall framework of UniDet-D is illustrated in Fig.~\ref{FIG:fig2} (a). In general, the proposed model is a dual-branch network structure, including the detection sub-network and the enhancement sub-network. In addition to the detection results, the output of UniDet-D also includes the image restoration results. Specifically, the degraded images are input into three different encoders for subsequent feature fusion. Specifically, the degradation representation encoder is first utilized to capture the multi-layer degradation information from the input low-quality images. These learned degradation representations are used to assist the object detection and image enhancement encoder branches. The degradation-guided fuse process enhances the UniDet-D’s adaptability to various degradations, thereby improving the model’s capability to extract features from blurred images. Then, the feature fusion is performed on the outputs of three different visual encoders. By integrating features from multiple visual encoders, the representation capacity is enhanced, which in turn effectively aggregates diverse semantic and structural information, thereby enabling more complete feature learning. 

Notably, as shown in the magnified view of the object detection encoder in Fig.~\ref{FIG:fig2} (b), a dynamic spectral perceive-select strategy is developed. This strategy aims to dynamically extract and emphasize informative spectral components while suppressing redundant frequencies that may interfere with degraded feature representations. Specifically, the MSP and $\mathrm{AF^{2}}$ modules are proposed to tackle the difficulty of insufficient frequency domain information utilization, thereby enhancing the robustness and precision of object detection in challenging weather. 
Finally, the fused features possess the information required for both object detection and image restoration tasks. They are fed into an image restoration decoder and an object detection regression module for rendering the final enhanced image and generating the detection results, respectively.

\subsection{Dynamic Spectral Perceive-Select Strategy}
\label{subsec2}
The dynamic spectral perceive-select design is motivated by the theoretical analysis of how vital visual details are lost in adverse-weather imagery.  Based on this insight, we further elaborate on the details of the proposed MSP and $\mathrm{AF^{2}}$ modules in the following Subsections.

\subsubsection{Image Frequency Utilization Analysis}

Channel attention is widely employed in Convolutional Neural Networks (CNNs), where it assigns importance to different feature channels using scalar weights. Typically, global average pooling (GAP) is adopted as a straightforward solution for scalar computation due to its simplicity and efficiency. Consider an image feature tensor $X \in {\mathbb{R}^{C \times H \times W}}$, where $C$ is the number of channels, and $H$ and $W$ denote the height and width of the feature map, respectively. The channel attention can be formulated as
\begin{equation}\label{1}
\mathrm{Att_{orig}} = \sigma\left( \mathrm{Fc}\left( \mathrm{GAP}(X) \right) \right),
\end{equation}
where $\mathrm{Att_{orig}} \in {\mathbb{R}^C}$ refers to the attention vector, $\sigma$ denotes the sigmoid activation function, and $\mathrm{Fc}$ denotes a mapping function, such as a fully connected layer. The $\mathrm{GAP}$ operation summarizes each channel into a scalar by averaging across its spatial dimensions. After calculating the attention vector, the feature channels of $X$ are re-weighted as
\begin{equation}
\hat{X}_{:, i, :, :} = \mathrm{Att}_{i} \cdot X_{:, i, :, :}, \quad \text{where } i \in \{0, 1, \dots, C{-}1\},
\end{equation}
where $\hat{X}$ denotes the output tensor after attention modulation.

Drawing from principles in digital signal processing, the DCT is used to analyze the frequency characteristics of feature maps. Specifically, the DCT basis function for spatial location $(i,j)$ under frequency $(h,w)$ is defined as
\begin{equation}
D_{h,w}^{i,j} = \cos\left( \frac{\pi h}{H} \left(i + \frac{1}{2} \right) \right) \cdot \cos\left( \frac{\pi w}{W} \left(j + \frac{1}{2} \right) \right),
\end{equation}
where $h \in \{0,\dots,H{-}1\}$ and $w \in \{0,\dots,W{-}1\}$. Given an input feature patch $x^{2d} \in \mathbb{R}^{H \times W}$, its 2D DCT spectrum $f^{2d} \in \mathbb{R}^{H \times W}$ is computed as
\begin{align}\label{DCT}
f_{h,w}^{2d} &= \sum_{i = 0}^{H - 1} \sum_{j = 0}^{W - 1} x_{i,j}^{2d} \cdot D_{h,w}^{i,j},
\end{align}
where $x_{i,j}^{2d}$ denotes the spatial value at $(i,j)$, and $f_{h,w}^{2d}$ is the frequency coefficient at $(h,w)$.

When $h = 0$ and $w = 0$, the lowest frequency component can be simplified as
\begin{align}
f_{0,0}^{2d} 
&= \sum_{i=0}^{H-1} \sum_{j=0}^{W-1} x_{i,j}^{2d} \nonumber \\
&= \mathrm{GAP}\left(x^{2d}\right) \cdot H \cdot W,
\end{align}
where $f_{0,0}^{2d}$ corresponds to the low-frequency component. This implies that GAP is essentially a scaled version of the low-frequency component of the 2D DCT. Therefore, DCT can be viewed as a generalization of GAP that preserves richer frequency characteristics.

Similarly, the inverse 2D DCT is given by
\begin{align}\label{inverse}
x_{i,j}^{2d} &= \sum_{h = 0}^{H - 1} \sum_{w = 0}^{W - 1} f_{h,w}^{2d} \cdot D_{h,w}^{i,j} \nonumber \\
&= f_{0,0}^{2d} D_{0,0}^{i,j} + \!  \!  \!  \sum_{\substack{(h,w) \neq (0,0)}} f_{h,w}^{2d} \cdot D_{h,w}^{i,j} \nonumber \\
&= \mathrm{GAP}(x^{2d}) \cdot HW \cdot D_{0,0}^{i,j} + \!  \!  \!  \!  \!   \sum_{\substack{(h,w) \neq (0,0)}} f_{h,w}^{2d} \cdot D_{h,w}^{i,j},
\end{align}
where $D_{0,0}^{i,j}$ is a constant value dependent on position $(i,j)$. Based on Eq. \eqref{inverse}, the original feature $x_{i,j}^{2d}$ is reconstructed from both low- and high-frequency components. However, in the attention mechanism defined by Eq. \eqref{1}, only the lowest frequency is retained. Therefore, the reconstructed tensor can be approximated as

\begin{equation}
\label{7}
\begin{aligned}
X =\underbrace{\mathrm{GAP}(x^{2d}) \cdot H \cdot W \cdot D_{0,0}^{i,j}}_{\text{utilized}}&+ \underbrace{\sum\nolimits_{(h,w)\neq(0,0)} f_{h,w}^{2d} \cdot D_{h,w}^{i,j}}_{\text{discarded}},
\end{aligned}
\end{equation}
which indicates that traditional GAP-based attention discards informative mid- and high-frequency signals, which are essential for robust object recognition under degradation.

Some recent works~\cite{qin2021fcanet,xie2023boosting} attempt to include additional frequency components by manually selecting a fixed subset of DCT channels. Let $\mathcal{I}_{\text{fixed}} \subseteq \{(h,w)\}$ denote such a predefined index set, the reconstructed feature becomes:
\begin{equation}
X_{\text{fixed}} = \sum_{(h,w)\in \mathcal{I}_{\text{fixed}}} f_{h,w}^{2d} \cdot D_{h,w}^{i,j}.
\end{equation}
However, this static strategy lacks adaptability. Because $\mathcal{I}_{\text{fixed}}$ is not dynamically adjusted according to image content or degradation type, many discriminative components may be ignored. As a result, fixed frequency selection introduces spectral bias and hinders generalization to unseen degradations.

\textbf{\textit{Insights Summary}}. Through our frequency-domain analysis, we identify two fundamental limitations in existing feature learning mechanisms: \textit{(i) GAP-induced frequency information discard}, which discards informative mid- and high-frequency details that are crucial for accurate object recognition under degradations. \textit{(ii) Fixed frequency selection limits generalization}, as such fixed selection schemes lack adaptability and introduce spectral bias, thereby degrading performance when encountering unseen or complex degradation patterns. To overcome these challenges, we propose two key improvements, which are elaborated in the following subsections.

\subsubsection{Multi-Spectrum Perception (MSP)}
To introduce more information and enhance feature representation, we propose extending GAP to incorporate additional frequency components from the 2D DCT and capturing more comprehensive information through multi-spectral components. The MSP procedure is shown in Fig.~\ref{FIG:fig2} (b). 

First, the image feature tensor $X \in \mathbb{R}^{C \times H \times W}$ is evenly partitioned along the channel dimension into $n$ segments. Let the resulting parts be denoted as $\left[ X^0, X^1, \dots, X^{n-1} \right]$, where $X^i \in \mathbb{R}^{C' \times H \times W}$ and $C' = \frac{C}{n}$. For each part $X^i$, a designated 2D DCT frequency component is assigned. The projection result at the $(u_i, v_i)$-th frequency location $\mathrm{Freq}^i \in \mathbb{R}^{C'}$ is used as a compressed representation:
\begin{align}
\mathrm{Freq}^i &= \mathrm{DCT}^{(u_i, v_i)}\left( X^i \right)  \nonumber\\
&= \sum_{h=0}^{H-1} \sum_{w=0}^{W-1} X^i_{:, h, w} \cdot D^{u_i, v_i}_{h, w}, i \in \{0, 1, \dots, n - 1\},
\end{align}
where $(u_i, v_i)$ denotes the 2D frequency index associated with the $i$-th segment and $D^{u_i,v_i}_{h,w}$ is the DCT basis function evaluated at spatial location $(h,w)$.

The complete multi-frequency vector is constructed by concatenating all partial vectors:
\begin{align}
\mathrm{Freq} &= \mathrm{DCT}^{2D}(X) \nonumber \\
&= \mathrm{Concat}\left( \mathrm{Freq}^0, \mathrm{Freq}^1, \dots, \mathrm{Freq}^{n-1} \right),
\end{align}
where $\mathrm{DCT}^{2D}(X)$ denotes the frequency-aware projection applied over all segments, and $\mathrm{Concat}(\cdot)$ indicates concatenation along the channel axis. The final vector $\mathrm{Freq} \in \mathbb{R}^C$ encodes multi-spectral responses across all channels and serves as the input to subsequent frequency-aware filtering modules.

\subsubsection{Adaptable Frequency Filtering ($AF^{2}$)}

Image features comprise various frequencies, yet not all frequency channels provide equally useful information for object detection. To mitigate noise disturbances and enhance spectrum utilization, we introduce a $\mathrm{AF^{2}}$ module that exploits the relative significance of each frequency component. This module uses learning-driven switches that assign a binary score to each frequency channel: salient channels are given a score of one, while others are assigned a score of zero. Channels that receive a score of zero are then filtered out from the model. This simple yet effective module reduces the computational burden of domain transformation. 

Fig.~\ref{FIG:fig2} (b) illustrates the pipeline of the $\mathrm{AF^{2}}$ module. The full set of frequency components is processed through a $1 \times 1$ convolutional layer with two kernels, producing a tensor named freqLogits of shape $\mathrm{C \times H \times W \times 2} $. This tensor represents the switching scores for each frequency component, where the last dimension models the binary selection probabilities via a softmax function.

\begin{figure}[t]
    \centering
    \includegraphics[width=1\linewidth]{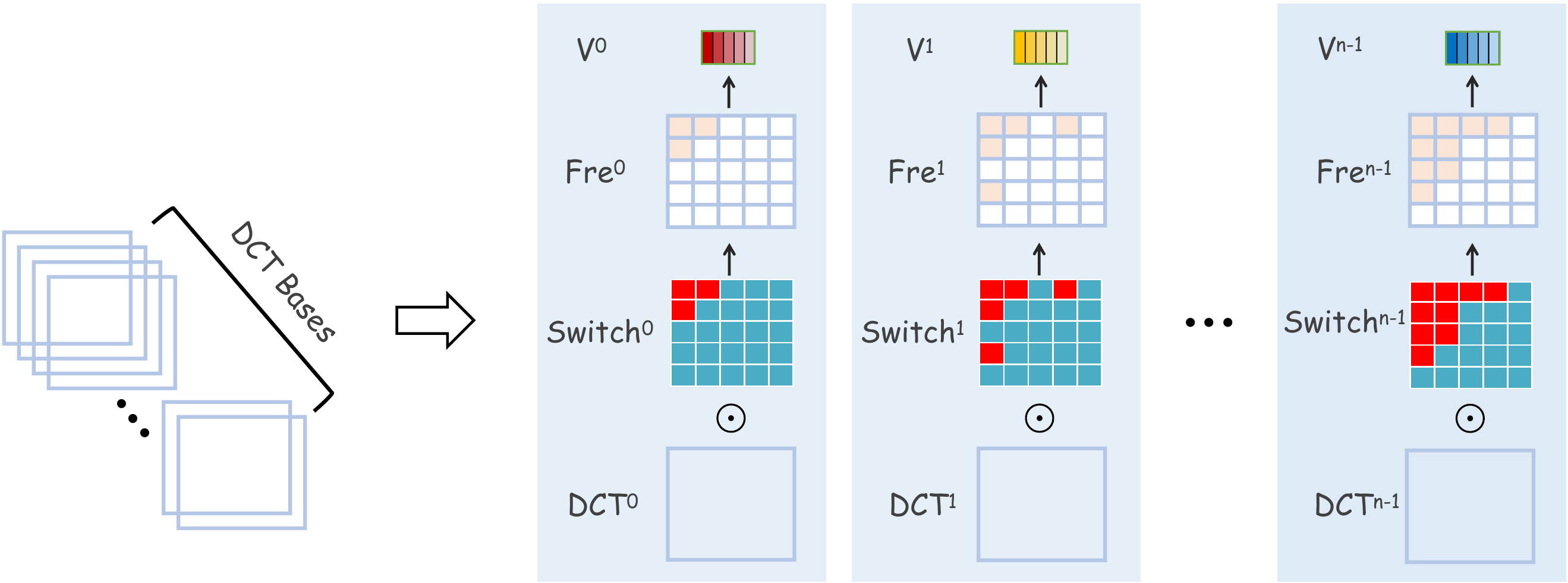}
    \caption{
    A more detailed visualization of the adaptable frequency filtering mechanism. Each spectral map is selectively filtered by learnable binary switches to produce task-relevant frequency representations.
Specifically, each frequency map (e.g., $\mathrm{DCT}^0, \mathrm{DCT}^1, \dots, \mathrm{DCT}^{n{-}1}$) is individually passed through a binary switch map ($\mathrm{Switch}^i$), which selectively retains or discards frequency components. The filtered output is then aggregated to form $\mathrm{Fre}^i$, from which a compact representation $V^i$ is generated via spectral encoding. The entire process is guided by learnable switching behavior for each frequency channel, enabling dynamic and data-adaptive spectrum selection.
    }
    \label{FIG:fig3}
\end{figure}

During inference, the two values in each freqLogits vector determine the likelihood of turning a frequency component “on” or “off.” For example, if the two values in a frequency channel are 0.8 and 0.2, there is an 80\% chance that this frequency will be suppressed. These sampled binary switches are then point-wise multiplied with the frequency component tensor to obtain the final selected spectrum.
However, as Bernoulli sampling is non-differentiable, we adopt the Gumbel-Softmax trick~\cite{jang2017categorical} to enable gradient backpropagation during training, making the frequency selection process learnable in an end-to-end manner. A more detailed visualization of the learnable frequency filtering mechanism is illustrated in Fig.~\ref{FIG:fig3}.

Let $f_{h,w}^{2d}$ denote the frequency component at position $(h, w)$ of the input feature map after 2D-DCT. The proposed $\mathrm{AF^{2}}$ module is denoted by $S$, where $S(f_{h,w}^{2d}) \in \{0, 1\}$. A frequency component is retained if
\begin{equation}
S(f_{h,w}^{2d}) \ne 0, \quad \text{i.e., } S(f_{h,w}^{2d}) \cdot f_{h,w}^{2d} \ne 0.
\end{equation}
To regularize the selection behavior and mitigate spectral bias, we introduce a frequency loss function $\mathcal{L}_{fre}$~\cite{zhang2023frequency}, which encourages sparsity in frequency selection as
\begin{equation}
\mathcal{L}_{fre} = \log \left( \sum_{h=0}^{H-1} \sum_{w=0}^{W-1} S(f_{h,w}^{2d}) \right),
\end{equation}
where the logarithmic function is used to smooth the binary selection values, addressing the stepped distribution characteristics and enhancing training stability.

This frequency-aware constraint is integrated into the overall training objective, which combines detection, restoration, contrastive, and frequency selection losses. The total loss is defined as
\begin{equation}
\mathcal{L}_{Total} = \lambda_1 \mathcal{L}_{Det} + \lambda_2 \mathcal{L}_{Res} + \lambda_3 \mathcal{L}_{Fre} + \lambda_4 \mathcal{L}_{Cls},
\end{equation}
where $\mathcal{L}_{Det}$ denotes the object detection loss, $\mathcal{L}_{Res}$ is the image restoration loss, $\mathcal{L}_{Fre}$ is the proposed frequency regularization term, and $\mathcal{L}_{Cls}$ refers to the contrastive loss. Following previous works~\cite{wang2024RDMNet, zhang2023frequency, wang2023restornet}, we set the hyperparameters as $\lambda_1 = 0.2$, $\lambda_2 = 0.8$, $\lambda_3 = 0.1$, and $\lambda_4 = 0.1$, respectively.

\begin{table*}[t]
\caption{Evaluation results on the VRain-test dataset.}
\label{table1}
\centering
\scriptsize
\resizebox{\textwidth}{!}{%
\begin{tabular}{lccccccc}
\hline
\multicolumn{1}{c}{Method}                  & Type       & Person & Bicycle & Car   & Motorbike & Bus   & mAP   \\ \hline
YOLOXs~\cite{ge2021yolox}                   & Baseline   & 80.98  & 65.88   & 74.54 & 70.83     & 84.35 & 75.32 \\
YOLOXs*~\cite{ge2021yolox}                  & Baseline   & 75.94  & 63.00   & 66.92 & 65.32     & 73.10 & 68.86 \\
AirNet-YOLOXs*~\cite{li2022AirNet}          & Restore    & 80.06  & 67.81   & 70.75 & 69.87     & 82.77 & 74.25 \\
RestorNet-YOLOXs*~\cite{wang2023restornet}  & Restore    & 80.44  & 68.29   & 71.68 & 70.16     & 82.54 & 74.62 \\
TogetherNet~\cite{wang2022togethernet}      & Multi-task & 83.15  & 70.31   & 77.18 & 73.08     & 82.93 & 77.33 \\
RDMNet~\cite{wang2024RDMNet}                & Multi-task & 82.97  & 70.17   & 77.82 & 76.36     & 84.74 & 78.41 \\
\textbf{UniDet-D}                           & Multi-task & \textbf{83.45} & \textbf{70.99} & \textbf{78.53} & \textbf{76.64} & \textbf{85.91} & \textbf{79.10} \\
\hline
\end{tabular}%
}
\begin{flushleft}
\scriptsize \textit{Note: Methods marked with an asterisk (*) are trained on clean VOC images, while others are trained on degraded images.} 
\end{flushleft}
\end{table*}

\begin{table*}[th]
\caption{Evaluation results on the VSnow-test dataset.}\label{table5}
\centering
\resizebox{\textwidth}{!}{%
\begin{tabular}{lccccccc}
\hline
\multicolumn{1}{c}{Method}                & Type       & Person & Bicycle        & Car   & Motorbike & Bus            & mAP   \\ \hline
YOLOXs                                    & Baseline   & 81.16  & 66.64          & 75.43 & 70.87     & 83.28          & 75.48 \\
YOLOXs*                                   & Baseline   & 78.40  & 64.88          & 70.8  & 56.90     & 81.15          & 70.43 \\
TKL-YOLOXs*~\cite{chen2022learning}       & Restore    & 81.02  & 70.14          & 75.53 & 70.33     & 84.42          & 76.29 \\
LMQFormer-YOLOXs*~\cite{lin2023lmqformer} & Restore    & 81.36  & 70.44 & 77.57 & 71.61     & 84.35          & 77.07 \\
TogetherNet                               & Multi-task & 82.20  & 69.7           & 77.72 & 71.39     & 85.30 & 77.26 \\
RDMNet                                    & Multi-task & 83.42  & 70.44 & 77.44 & 72.11     & 84.78          & 77.63 \\
{\color[HTML]{000000} \textbf{UniDet-D}} &
  {\color[HTML]{000000} Multi-task} &
  {\color[HTML]{000000} \textbf{84.52}} &
  {\color[HTML]{000000} \textbf{70.47}} &
  {\color[HTML]{000000} \textbf{78.73}} &
  {\color[HTML]{000000} \textbf{73.35}} &
  {\color[HTML]{000000} \textbf{85.91}} &
  {\color[HTML]{000000} \textbf{78.58}} \\ \hline
\end{tabular}%
}
\end{table*}

\begin{table*}[t]  
\caption{Evaluation results on the VFog-test dataset.}
\label{table2}
\centering
\resizebox{\textwidth}{!}{%
\begin{tabular}{lccccccc}
\hline
\multicolumn{1}{c}{Method}                 & Type           & Person         & Bicycle        & Car            & Motorbike & Bus   & mAP   \\ \hline
YOLOXs                                     & Baseline       & 67.67          & \textbf{83.28} & 77.75          & 68.91     & 81.70 & 75.86 \\
YOLOXs*                                    & Baseline       & 73.09          & 57.22          & 69.55          & 59.83     & 77.34 & 67.41 \\
DCP-YOLOXs*~\cite{he2010DCP}               & Restore        & 81.84          & 70.38          & 78.63          & 73.48     & 64.68 & 77.80 \\
AODNet-YOLOXs*~\cite{li2017aod}            & Restore        & 67.40          & 49.19          & 60.51          & 55.59     & 62.07 & 58.95 \\
AECRNet-YOLOXs*~\cite{wu2021contrastive}   & Restore        & 80.47          & 67.82          & 76.97          & 72.46     & 82.73 & 76.09 \\
RestorNet-YOLOXs*                          & Restore        & 78.71          & 67.15          & 72.56          & 71.68     & 82.36 & 74.46 \\
DS-Net~\cite{huang2021dsnet}               & Multi-task     & 72.44          & 60.47          & 81.27          & 53.85     & 61.43 & 65.89 \\
IA-YOLO~\cite{liu2022IA-YOLO}              & Image adaptive & 70.98          & 61.98          & 70.98          & 57.93     & 61.98 & 64.77 \\
TogetherNet                                & Multi-task     & \textbf{84.11} & 69.26          & 79.59          & 72.12     & 85.62 & 78.17 \\
RDMNet                                     & Multi-task     & 83.58          & 69.61          & 79.95 & 74.59     & 85.23 & 78.59 \\

\textbf{UniDet-D} & Multi-task     & 83.98          & 73.56          & \textbf{79.98}          & \textbf{76.20} & \textbf{86.86} & \textbf{80.10} \\ \hline
\end{tabular}%
}
\end{table*}

\begin{table*}[htb]
\caption{Evaluation results under various weather conditions.}\label{various}
\centering
\resizebox{\textwidth}{!}{%
\begin{tabular}{lcccccc}
\hline
\multicolumn{1}{c}{Method} & Type       & VOC-test & VRain-test & VFog-test & VSnow-test & Average \\ \hline
YOLOXs                     & Baseline   & 75.77    & 74.50         & 75.50        & 74.24         & 75.00   \\
YOLOXs*                    & Baseline   & 79.09    & 68.86         & 67.41        & 70.43         & 71.45   \\
TogetherNet                & Multi-task & 78.86    & 76.74         & 76.78        & 77.03         & 77.35   \\
RDMNet                     & Multi-task & 79.01    & 77.23         & 76.89        & 77.18         & 77.58   \\

{\color[HTML]{000000} \textbf{UniDet-D}} &
  {\color[HTML]{000000} Multi-task} &
  {\color[HTML]{000000} \textbf{79.95}} &
  {\color[HTML]{000000} \textbf{78.67}} &
  {\color[HTML]{000000} \textbf{78.87}} &
  {\color[HTML]{000000} \textbf{77.54}} &
  {\color[HTML]{000000} \textbf{78.76}} \\ \hline
\end{tabular}%
}
\end{table*}

\begin{table*}[!h]
\caption{Evaluation results on the \textbf{unseen} RxDark dataset (\textbf{Real-world Data}).}
\label{zero}
\centering
\resizebox{\textwidth}{!}{%
\begin{tabular}{lccccccc}
\hline
Method                  & Type           & Person         & Bicycle        & Car            & Motorbike      & Bus            & mAP            \\ \hline
YOLOXs                  & Baseline       & 52.72          & \textbf{66.01} & 56.34          & 37.27          & 55.13          & 53.49          \\
YOLOXs*                 & Baseline       & 45.51          & 61.25          & 50.63          & 33.12          & 49.68          & 48.04          \\
ZeroDCE-YOLOXs*~\cite{guo2020zero}         & Restore        & 55.12          & 63.65          & 58.16          & 42.38          & 51.24          & 54.11          \\
IAT-YOLOXs*~\cite{cui2022you}             & Restore        & 56.27          & 64.14          & 59.29          & 43.15          & 53.55          & 55.28          \\
IA-YOLO                 & Image adaptive & 45.49          & 55.20          & 46.39          & 20.45          & 52.72          & 44.05          \\
TogetherNet             & Multi-task     & 63.86          & 56.80          & 59.41          & 42.25          & 65.75          & 57.61          \\
RDMNet                  & Multi-task     & 64.51          & 59.17          & 57.71          & 46.37          & 69.65          & 59.48          \\
\textbf{UniDet-D}       & Multi-task     & \textbf{65.96} & 60.12 & \textbf{61.13} & \textbf{46.90} & \textbf{72.57} & \textbf{61.33} \\
\textbf{UniDet-D (T)}    & Multi-task     & \textbf{68.07} & \textbf{67.40} & \textbf{64.49} & \textbf{54.54} & \textbf{73.27} & \textbf{65.55} \\
\textit{Gain from training} & --         & \textcolor{blue}{+2.11} & \textcolor{blue}{+7.28} & \textcolor{blue}{+ 3.36} & \textcolor{blue}{+7.64} & \textcolor{blue}{+0.70} & \textcolor{blue}{+4.22} \\
\hline
\end{tabular}%
}
\begin{flushleft}
\scriptsize \textit{Note: UniDet-D denotes the model trained without low-light data. UniDet-D (T) is trained on synthetic low-light images, but has not seen real-world data from the ExDark.}
\end{flushleft}
\end{table*}

\section{Experiments and Analysis}
\label{sec4}

\subsection{Implementation Details}
\label{subsec3}
UniDet-D are trained using the SGD optimizer with an initial learning rate of $1 \times {10^{ - 2}}$. The total number of train epochs is set to 100, with a batch size of 16. A cosine annealing strategy is employed to adjust the learning rate dynamically. Both the training and testing images are resized to $640 \times{640}$. Unlike conventional object detection methods, we leverage paired low-quality and high-quality images to train the enhancement branch's feature extraction capability, thereby improving the performance of the object detection branch. To evaluate the performance of the proposed UniDet-D under adverse weather conditions, we adopt the mean average precision (mAP), a widely used objective evaluation metric in object detection tasks.

\subsection{Adverse-weather Dataset Construction}
To evaluate UniDet-D under diverse adverse weather conditions, we construct or adopt five degraded object detection datasets derived from the PASCAL VOC dataset~\cite{Mark2010VOC}, each containing five object categories: car, bus, motorbike, bicycle, and person.

\textit{Rainy Weather (VRain):}
We generate a synthetic rainy dataset, VRain, by applying the rain streak synthesis method from RainDS~\cite{quan2021removing} to VOC images. The resulting dataset contains 9,578 training images (VRain-train) and 2,129 test images (VRain-test), featuring rain streaks of varying intensity and orientation. 

\textit{Foggy Weather (VFog):}
We adopt the synthetic hazy dataset VFog~\cite{wang2022togethernet}, which is generated by applying an atmospheric scattering model to the clean images from the VOC dataset. The dataset is divided into a training set of 9,578 images (VFog-train) and a test set of 2,129 images (VFog-test).

\textit{Snowy Weather (VSnow):}
We synthesize the VSnow dataset using snow masks from the CSD dataset~\cite{chen2021CSD}, applied with varying intensity (weights between 0.5 and 1.0). This results in 9,578 snow-affected training images (VSnow-train) and 2,219 test images (VSnow-test).

\textit{{Low-light Condition (VLow Light):}} 
To simulate low-light conditions for training purposes, we construct the VLow Light dataset by applying a pixel-wise transformation $f(x) = x^r$ to the VOC dataset, where $r$ is randomly sampled from the range $[1.5, 5]$. This process generates images with varying degrees of darkness, providing diverse low-light samples to enhance the model’s robustness during training.

\textit{Mixed Weather Dataset:}
To train a unified model, we construct a mixed training dataset by combining the VOC, VRain, VFog, and VSnow datasets, yielding 38,312 images in total. The model is evaluated on the corresponding test sets across all weather scenarios.
\begin{figure*}[t]
    \centering
    \includegraphics[width=0.875\linewidth]{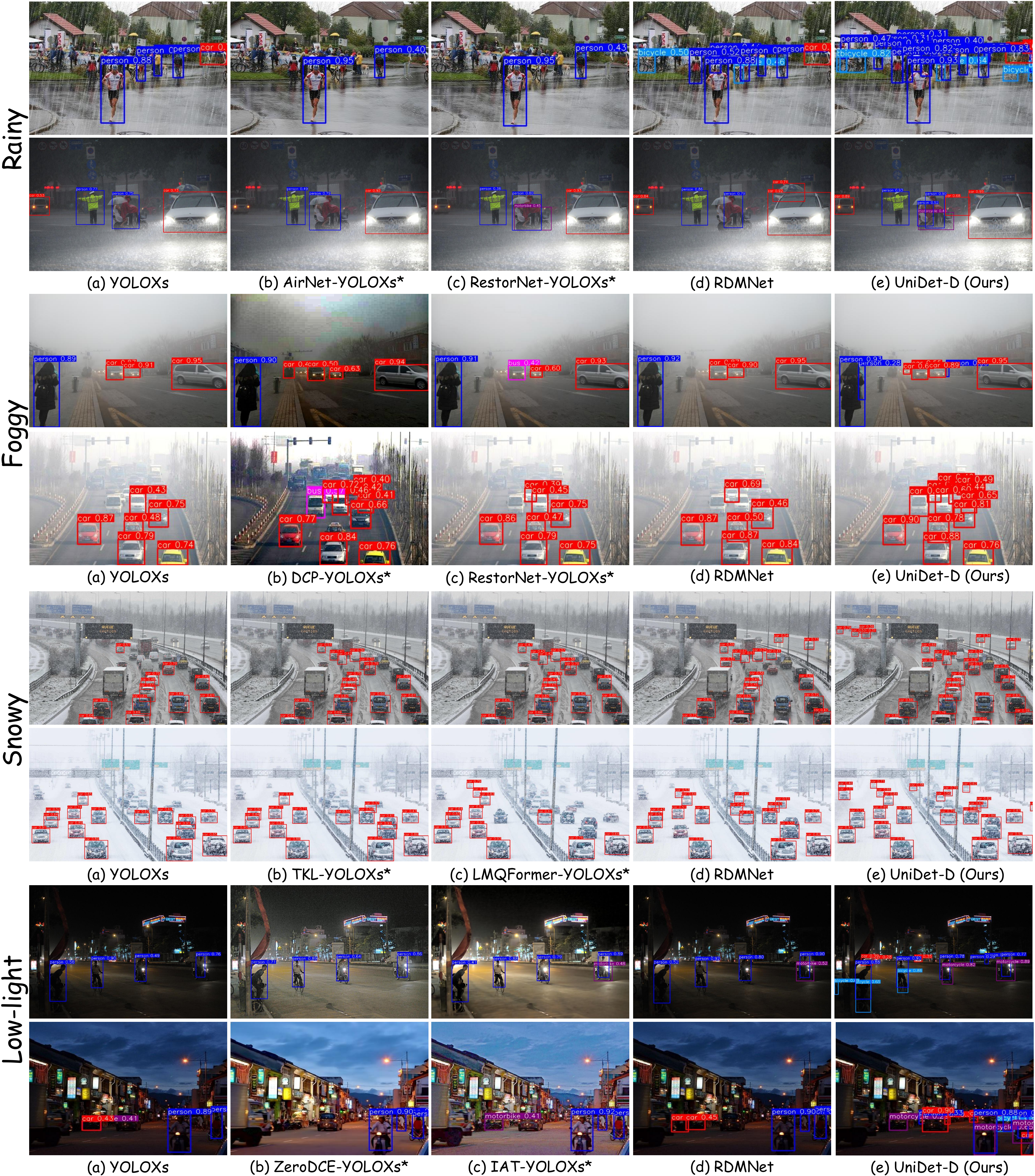}
    \caption{
Visualization of detection results under different weather degradations (synthetic and real-world). UniDet-D consistently achieves more accurate detections with reduced false positives and missed targets compared to existing SOTA methods. \textit{\textbf{Note}: Bounding boxes are color-coded by object category: red for vehicles, dark blue for pedestrians, light blue for bicycles, pink for buses, and purple for motorbikes.}
}
    \label{FIG:all}
\end{figure*}

\subsection{Comparative Experiments}

\subsubsection{Experiments on Rainy Weather}

To evaluate UniDet-D under rainy conditions, we use the synthetic VRain dataset to compare its performance with SOTA multi-task and ``restore + detect'' methods. All models are trained on VRain-train and tested on VRain-test. YOLOXs~\cite{ge2021yolox} is used as the baseline, with models trained on rainy and clean images denoted as YOLOXs and YOLOXs*, respectively. For ``restore + detect,'' we apply AirNet~\cite{li2022AirNet} and RestorNet~\cite{wang2023restornet} as preprocessing before YOLOXs*, forming AirNet-YOLOXs* and RestorNet-YOLOXs*. For multi-task comparison, we include TogetherNet~\cite{wang2022togethernet} and RDMNet~\cite{wang2024RDMNet}. As shown in Table~\ref{table1}, UniDet-D outperforms all compared methods on VRain-test, achieving a 1.08\% mAP improvement over RDMNet. 

\subsubsection{Experiments on Snowy Weather}

We evaluate UniDet-D on the synthetic VSnow dataset, comparing it with YOLOXs, YOLOXs*, and ``restore + detect'' methods using TKL~\cite{chen2022learning} and LMQFormer~\cite{lin2023lmqformer}. These snow removal models are combined with YOLOXs* to form TKL-YOLOXs* and LMQFormer-YOLOXs*. Multi-task baselines include TogetherNet and RDMNet. As shown in Table~\ref{table5}, UniDet-D achieves superior performance on VSnow-test, surpassing TogetherNet and RDMNet by 0.95\% and 1.32\% in mAP, respectively. These results confirm UniDet-D's robustness in detecting densely distributed and distant objects under snowy conditions.

\subsubsection{Experiments on Foggy Weather}

We compare UniDet-D with representative 'restore + detect' and end-to-end methods. ``restore + detect'' methods apply image dehazing (DCP~\cite{he2010DCP}, AODNet~\cite{li2017aod}, AECRNet~\cite{wu2021contrastive}, and RestorNet) followed by detection using YOLOXs* trained on clean images. These dehazing models are trained on the OTS subset of RESIDE~\cite{li2018RTTS} to align with RTTS data characteristics. Combined models include DCP-YOLOXs*, AODNet-YOLOXs*, etc.
End-to-end methods directly perform detection on foggy inputs, including DS-Net~\cite{huang2021dsnet}, TogetherNet, and IA-YOLO~\cite{liu2022IA-YOLO}. As shown in Table~\ref{table2}, UniDet-D outperforms all baselines on the VFog-test dataset, with a 1.51\% mAP gain over the recent RDMNet.

\subsubsection{Experiments under Various Degradations}
\label{subsec6}

To comprehensively assess the generalization capability of UniDet-D, we evaluate its performance across multiple weather conditions, including clean, rainy, snowy, and foggy environments. UniDet-D is trained on a mixed dataset comprising all four conditions. Baselines include YOLOXs and YOLOXs*, trained separately on corrupted and clean images, as well as multi-task models, TogetherNet and RDMNet.
As shown in Table~\ref{various}, UniDet-D consistently outperforms all compared methods, with mAP gains of 1.18\% and 1.41\% over TogetherNet and RDMNet, respectively. These results confirm the effectiveness of UniDet-D as a unified model capable of robust object detection across diverse degradations.

\subsubsection{Generalization Performance Evaluation}

To validate the generalization capability of the proposed UniDet-D model, this study uses the ExDark dataset~\cite{Exdark}, which consists of real-world low-light images and serves as an unseen test set. The evaluation compares UniDet-D with representative "restore + detect" and end-to-end baselines. The "restore + detect" approaches include ZeroDCE-YOLOXs~\cite{guo2020zero} and IAT-YOLOXs~\cite{cui2022you}. In addition, we also compare against the image-adaptive method IA-YOLO, as well as multi-task learning models such as TogetherNet and RDMNet, which are representative end-to-end solutions for degraded object detection. As shown in Table~\ref{zero}, under the challenging zero-shot setting, UniDet-D demonstrates superior generalization, outperforming RDMNet by 1.85 mAP and achieving notable AP gains of 3.42 on car and 2.92 on bus. Furthermore, when training on synthetic low-light data VLow Light, we denote as UniDet-D (T), which improves its mAP by 4.22 $\%$, demonstrating both robust adaptability and the effectiveness of its unified design.

\subsection{Qualitative Evaluation}
\begin{figure}[t]
    \centering    \includegraphics[width=1\linewidth]{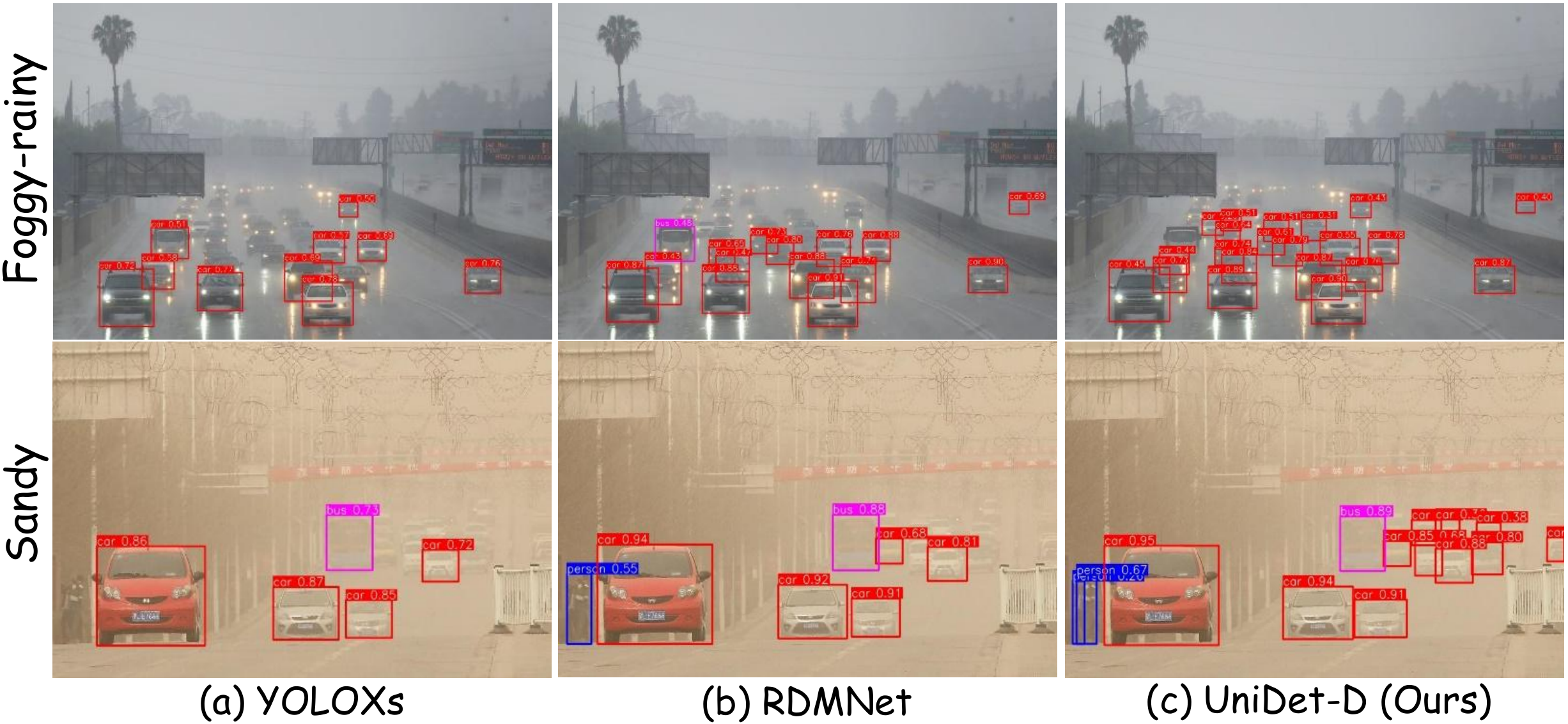}
    \caption{
Visualization of detection results under unseen weather degradations, including sandy and rain-fog mixed weather. \textit{\textbf{Note}: Bounding boxes are color-coded by object category: red for vehicles, dark blue for pedestrians, and purple for motorbikes.}
}
    \label{FIG:G}
\end{figure}
\subsubsection{Comprehensive Comparison under Various Degradations}
In addition to quantitative evaluation, we present a comprehensive qualitative study across diverse adverse weather conditions, including rain, snow, fog, and low-light scenarios, as shown in Fig.~\ref{FIG:all}. The evaluated models include the baseline YOLOXs, enhancement-based pipelines such as AirNet-YOLOXs, DCP-YOLOXs*, RestorNet-YOLOXs*, ZeroDCE-YOLOXs*, IAT-YOLOXs*, TKL-YOLOXs*, and LMQFormer-YOLOXs*, as well as the recent multi-task model RDMNet. Notably, UniDet-D consistently achieves clearer object localization with fewer missed or false detections compared to these state-of-the-art approaches. In contrast, existing models exhibit detection failures under severe degradations such as dense fog or nighttime conditions.

\subsubsection{Generalization Comparison under Unseen Degradations}
To assess the generalization ability of UniDet-D, we further evaluate it under previously unseen degradation types, including sandstorms and combined rain-fog scenarios, as illustrated in Fig.~\ref{FIG:G}. While the baseline detector and strong multi-task model like RDMNet struggle in these complex and untrained conditions, UniDet-D effectively identifying and localizing more targets. These results highlight the adaptability and strong generalization capability of UniDet-D, largely attributed to the proposed dynamic spectral learning.

\subsection{Ablation Studies}
To evaluate the impact of the frequency filtering switches, we conduct the ablation experiment, with the results presented in Table~\ref{table9}. All ablation models are trained on the VFog-train data and tested on the VFog-test data. The DCT of the multi-spectral attention mechanism and the learning-driven frequency filtering are progressively introduced into UniDet-D. A fascinating finding is that UniDet-D trained on a subset of the frequency components achieves better performance than that trained on all frequency bands, with a 1.54 $\%$ mAP improvement. This result suggests that not all frequency components contribute equally to detection accuracy, and selectively retaining informative frequencies can effectively suppress redundant or noisy signals. Such selective emphasis leads to more compact and robust feature representations, ultimately enhancing object detection performance.

\begin{table}[t]
\caption{Ablation results on frequency switches.}
\label{table9}
\centering
\setlength{\tabcolsep}{6pt} 
\begin{tabular}{l@{\hskip 7pt}c@{\hskip 6pt}c@{\hskip 7pt}c@{\hskip 7pt}c@{\hskip 6pt}c@{\hskip 7pt}c@{\hskip 7pt}c}
\hline
DCT & Switches & Person & Bicycle & Car & Motorbike & Bus & mAP \\
\hline
\checkmark & $\times$   & 82.06  & 70.45   & 79.38 & 74.50     & 86.42 & 78.56 \\
\checkmark & \checkmark & \textbf{83.98}  & \textbf{73.56}   & \textbf{79.92} & \textbf{76.20}     & \textbf{86.86} & \textbf{80.10} \\
\hline
\end{tabular}
\end{table}

\section{Conclusion}
In this paper, we propose UniDet-D, a unified object detection framework for low-quality imagery under diverse adverse weather conditions. Unlike existing approaches that are tailored to specific degradation types, UniDet-D jointly performs object detection and image restoration within a single end-to-end architecture. Motivated by a frequency-domain analysis of degraded images, we introduce a dynamic frequency attention mechanism comprising an MSP module and a $\mathrm{AF^{2}}$ method, enabling the model to adaptively select informative spectral components. Extensive experiments demonstrate that UniDet-D consistently outperforms SOTA detectors across a range of degraded scenarios, while exhibiting strong generalization to unseen conditions such as sandstorms and rain-fog mixtures. These results confirm the effectiveness and scalability of our approach, highlighting its potential for real-world deployment in low-quality visual perception tasks.

\bibliographystyle{IEEEtran} 
\bibliography{my}            

\begin{thebibliography}{10}
\providecommand{\url}[1]{#1}
\csname url@samestyle\endcsname
\providecommand{\newblock}{\relax}
\providecommand{\bibinfo}[2]{#2}
\providecommand{\BIBentrySTDinterwordspacing}{\spaceskip=0pt\relax}
\providecommand{\BIBentryALTinterwordstretchfactor}{4}
\providecommand{\BIBentryALTinterwordspacing}{\spaceskip=\fontdimen2\font plus
\BIBentryALTinterwordstretchfactor\fontdimen3\font minus \fontdimen4\font\relax}
\providecommand{\BIBforeignlanguage}[2]{{%
\expandafter\ifx\csname l@#1\endcsname\relax
\typeout{** WARNING: IEEEtran.bst: No hyphenation pattern has been}%
\typeout{** loaded for the language `#1'. Using the pattern for}%
\typeout{** the default language instead.}%
\else
\language=\csname l@#1\endcsname
\fi
#2}}
\providecommand{\BIBdecl}{\relax}
\BIBdecl

\bibitem{ge2021yolox}
Z.~Ge, S.~Liu, F.~Wang \emph{et~al.}, ``Yolox: Exceeding yolo series in 2021,'' \emph{arXiv preprint arXiv:2107.08430}, 2021.

\bibitem{zhang2023perception}
Y.~Zhang, A.~Carballo, H.~Yang \emph{et~al.}, ``Perception and sensing for autonomous vehicles under adverse weather conditions: A survey,'' \emph{ISPRS J. Photogramm. Remote Sens.}, vol. 196, pp. 146--177, 2023.

\bibitem{lee2022learning}
Y.~Lee, Y.~Kim, J.~Yu \emph{et~al.}, ``Learning to remove bad weather: Towards robust visual perception for self-driving,'' \emph{IEEE Rob. Autom. Lett.}, pp. 1--1, 2022.

\bibitem{zou2023object}
Z.~Zou, K.~Chen, Z.~Shi \emph{et~al.}, ``Object detection in 20 years: A survey,'' \emph{P IEEE}, vol. 111, no.~3, pp. 257--276, 2023.

\bibitem{zhang2024earthmarker}
W.~Zhang, M.~Cai, T.~Zhang, Y.~Zhuang, J.~Li, and X.~Mao, ``{EarthMarker}: A visual prompting multi-modal large language model for remote sensing,'' \emph{IEEE Trans. Geosci. Remote Sens.}, vol.~63, pp. 1--19, Jan. 2025.

\bibitem{zhang2024earthgpt}
W.~Zhang, M.~Cai, T.~Zhang, Y.~Zhuang, and X.~Mao, ``Earth{GPT}: A universal multi-modal large language model for multi-sensor image comprehension in remote sensing domain,'' \emph{IEEE Trans. Geosci. Remote Sens.}, vol.~62, pp. 1--20, Jun. 2024.

\bibitem{gao2022adaMixer}
Z.~Gao, L.~Wang, B.~Han \emph{et~al.}, ``Adamixer: A fast-converging query-based object detector,'' in \emph{Proc. IEEE/CVF Conf. Comput. Vis. Pattern Recognit.}, June 2022, pp. 5364--5373.

\bibitem{liu2022IA-YOLO}
W.~Liu, G.~Ren, R.~Yu \emph{et~al.}, ``Image-adaptive yolo for object detection in adverse weather conditions,'' in \emph{AAAI Conf. Artif. Intell.}, vol.~36, no.~2, 2022, pp. 1792--1800.

\bibitem{liu2023Improving}
W.~Liu, W.~Li, J.~Zhu \emph{et~al.}, ``Improving nighttime driving-scene segmentation via dual image-adaptive learnable filters,'' \emph{IEEE Trans. Circuits Syst. Video Technol.}, vol.~33, no.~10, pp. 5855--5867, 2023.

\bibitem{Kalwar2023GDIP}
S.~Kalwar, D.~Patel, A.~Aanegola \emph{et~al.}, ``Gdip: Gated differentiable image processing for object detection in adverse conditions,'' in \emph{IEEE Int. Conf. Robot. Autom.}, 2023, pp. 7083--7089.

\bibitem{xi2024detection}
Y.~Xi, W.~Jia, Q.~Miao \emph{et~al.}, ``Detection-driven exposure-correction network for nighttime drone-view object detection,'' \emph{IEEE Trans. Geosci. Remote Sens.}, 2024.

\bibitem{zhang2024cpa}
Y.~Zhang, Y.~Wu, Y.~Liu \emph{et~al.}, ``Cpa-enhancer: Chain-of-thought prompted adaptive enhancer for object detection under unknown degradations,'' \emph{arXiv preprint arXiv:2403.11220}, 2024.

\bibitem{oza2024Unsupervised}
P.~Oza, V.~A. Sindagi, V.~VS \emph{et~al.}, ``Unsupervised domain adaptation of object detectors: A survey,'' \emph{IEEE Trans. Pattern Anal. Mach. Intell.}, vol.~46, no.~6, pp. 4018--4040, 2024.

\bibitem{Li2022Cross}
G.~Li, Z.~Ji, X.~Qu \emph{et~al.}, ``Cross-domain object detection for autonomous driving: A stepwise domain adaptative yolo approach,'' \emph{IEEE Trans. Intell. Veh.}, vol.~7, no.~3, pp. 603--615, 2022.

\bibitem{wang2022togethernet}
Y.~Wang, X.~Yan, K.~Zhang \emph{et~al.}, ``Togethernet: Bridging image restoration and object detection together via dynamic enhancement learning,'' \emph{Comput. Graphics Forum}, vol.~41, no.~7, pp. 465--476, 2022.

\bibitem{huang2021dsnet}
S.-C. Huang, T.-H. Le, and D.-W. Jaw, ``Dsnet: Joint semantic learning for object detection in inclement weather conditions,'' \emph{IEEE Trans. Pattern Anal. Mach. Intell.}, vol.~43, no.~8, pp. 2623--2633, 2021.

\bibitem{zhong2024dehazing}
F.~Zhong, W.~Shen, H.~Yu \emph{et~al.}, ``Dehazing \& reasoning yolo: Prior knowledge-guided network for object detection in foggy weather,'' \emph{Pattern Recognit.}, vol. 156, p. 110756, 2024.

\bibitem{wang2024mdd}
N.~Wang, Y.~Wang, Y.~Feng \emph{et~al.}, ``Mdd-shipnet: Math-data integrated defogging for fog-occlusion ship detection,'' \emph{IEEE Trans. Intell. Transp. Syst.}, 2024.

\bibitem{qin2021fcanet}
Z.~Qin, P.~Zhang, F.~Wu \emph{et~al.}, ``Fcanet: Frequency channel attention networks,'' in \emph{Proc. IEEE/CVF Int. Conf. Comput. Vis.}, 2021, pp. 783--792.

\bibitem{xie2023boosting}
Z.~Xie, S.~Wang, K.~Xu \emph{et~al.}, ``Boosting night-time scene parsing with learnable frequency,'' \emph{IEEE Trans. Image Process.}, vol.~32, pp. 2386--2398, 2023.

\bibitem{ahmed1974discrete}
N.~Ahmed, T.~Natarajan, and K.~R. Rao, ``Discrete cosine transform,'' \emph{IEEE Trans. Comput.}, vol. 100, no.~1, pp. 90--93, 1974.

\bibitem{pang2019libra}
J.~Pang, K.~Chen, J.~Shi \emph{et~al.}, ``Libra r-cnn: Towards balanced learning for object detection,'' in \emph{Proc. IEEE Conf. Comput. Vis. Pattern Recog.}, 2019, pp. 821--830.

\bibitem{zhang2020dynamic}
H.~Zhang, H.~Chang, B.~Ma \emph{et~al.}, ``Dynamic r-cnn: Towards high quality object detection via dynamic training,'' in \emph{Proc. Eur. Conf. Comput. Vis.}\hskip 1em plus 0.5em minus 0.4em\relax Springer, 2020, pp. 260--275.

\bibitem{uijlings2013selective}
J.~R. Uijlings, K.~E. Van De~Sande \emph{et~al.}, ``Selective search for object recognition,'' \emph{Int. J. Comput. Vis.}, vol. 104, pp. 154--171, 2013.

\bibitem{liu2016ssd}
W.~Liu, D.~Anguelov, D.~Erhan \emph{et~al.}, ``Ssd: Single shot multibox detector,'' in \emph{Proc. Eur. Conf. Comput. Vis.}\hskip 1em plus 0.5em minus 0.4em\relax Springer, 2016, pp. 21--37.

\bibitem{Lin2017focal}
T.-Y. Lin, P.~Goyal, R.~Girshick \emph{et~al.}, ``Focal loss for dense object detection,'' in \emph{IEEE Conf. Comput. Vis. Pattern Recognit.}, 2017, pp. 2980--2988.

\bibitem{zhou2019objects}
X.~Zhou, D.~Wang, and P.~Kr{\"a}henb{\"u}hl, ``Objects as points,'' \emph{arXiv preprint arXiv:1904.07850}, 2019.

\bibitem{wang2023yolov7}
C.-Y. Wang, A.~Bochkovskiy, and H.-Y.~M. Liao, ``Yolov7: Trainable bag-of-freebies sets new state-of-the-art for real-time object detectors,'' in \emph{Proc. IEEE Conf. Comput. Vis. Pattern Recog. (CVPR)}, June 2023, pp. 7464--7475.

\bibitem{wang2025yolov9}
C.-Y. Wang, I.-H. Yeh, and H.-Y. Mark~Liao, ``Yolov9: Learning what you want to learn using programmable gradient information,'' in \emph{Proc. Eur. Conf. Comput. Vis.}\hskip 1em plus 0.5em minus 0.4em\relax Springer, 2025, pp. 1--21.

\bibitem{wang2024yolov10}
A.~Wang, H.~Chen, L.~Liu \emph{et~al.}, ``Yolov10: Real-time end-to-end object detection,'' \emph{arXiv preprint arXiv:2405.14458}, 2024.

\bibitem{zhang2022spectral}
J.~Zhang, J.~Huang, Z.~Tian \emph{et~al.}, ``Spectral unsupervised domain adaptation for visual recognition,'' in \emph{Proc. IEEE Conf. Comput. Vis. Pattern Recog.}, 2022, pp. 9829--9840.

\bibitem{li2023domain}
J.~Li, R.~Xu, J.~Ma \emph{et~al.}, ``Domain adaptive object detection for autonomous driving under foggy weather,'' in \emph{Proc. IEEE Winter Conf. Appl. Comput. Vis.}, January 2023, pp. 612--622.

\bibitem{kennerley2023pcnet}
M.~Kennerley, J.-G. Wang, B.~Veeravalli, and R.~T. Tan, ``2pcnet: Two-phase consistency training for day-to-night unsupervised domain adaptive object detection,'' in \emph{Proc. IEEE/CVF Conf. Comput. Vis. Pattern Recognit.}, June 2023, pp. 11\,484--11\,493.

\bibitem{wang2023ryolo}
L.~Wang, H.~Qin, X.~Zhou \emph{et~al.}, ``R-yolo: A robust object detector in adverse weather,'' \emph{IEEE Trans. Instrum. Meas.}, vol.~72, pp. 1--11, 2023.

\bibitem{pang2024mcnet}
J.~Pang, ``Mcnet: Magnitude consistency network for domain adaptive object detection under inclement environments,'' \emph{Pattern Recognit.}, vol. 145, p. 109947, 2024.

\bibitem{huang2021fsdr}
J.~Huang, D.~Guan, A.~Xiao, and S.~Lu, ``Fsdr: Frequency space domain randomization for domain generalization,'' in \emph{Proc. IEEE Conf. Comput. Vis. Pattern Recog.}, 2021, pp. 6891--6902.

\bibitem{cui2021multitask}
Z.~Cui, G.-J. Qi, L.~Gu \emph{et~al.}, ``Multitask aet with orthogonal tangent regularity for dark object detection,'' in \emph{Proceedings of the Proc. IEEE Int. Conf. Comput. Vis.}, October 2021, pp. 2553--2562.

\bibitem{xi2023coderainnet}
Y.~Xi, W.~Jia, Q.~Miao \emph{et~al.}, ``Coderainnet: Collaborative deraining network for drone-view object detection in rainy weather conditions,'' \emph{Remote Sens.}, vol.~15, no.~6, p. 1487, 2023.

\bibitem{xiao20233d}
A.~Xiao, J.~Huang, W.~Xuan, R.~Ren, K.~Liu, D.~Guan, A.~El~Saddik, S.~Lu, and E.~P. Xing, ``3d semantic segmentation in the wild: Learning generalized models for adverse-condition point clouds,'' in \emph{Proc. IEEE Conf. Comput. Vis. Pattern Recog.}, 2023, pp. 9382--9392.

\bibitem{wang2024RDMNet}
X.~Wang, X.~Liu, H.~Yang \emph{et~al.}, ``Degradation modeling for restoration-enhanced object detection in adverse weather scenes,'' \emph{IEEE Trans. Intell. Veh.}, 2024.

\bibitem{jang2017categorical}
E.~Jang, S.~Gu, and B.~Poole, ``Categorical reparametrization with gumble-softmax,'' in \emph{Proc. Int. Conf. Learn. Represent.}\hskip 1em plus 0.5em minus 0.4em\relax OpenReview. net, 2017.

\bibitem{zhang2023frequency}
D.~Zhang, Y.~Lu, Y.~Li \emph{et~al.}, ``Frequency learning attention networks based on deep learning for automatic modulation classification in wireless communication,'' \emph{Pattern Recognit.}, vol. 137, p. 109345, 2023.

\bibitem{wang2023restornet}
X.~Wang, H.~Chen, H.~Gou \emph{et~al.}, ``Restornet: An efficient network for multiple degradation image restoration,'' \emph{Knowledge-Based Systems}, vol. 282, p. 111116, 2023.

\bibitem{li2022AirNet}
B.~Li, X.~Liu, P.~Hu \emph{et~al.}, ``All-in-one image restoration for unknown corruption,'' in \emph{Proc. IEEE Conf. Comput. Vis. Pattern Recog.}, 2022, pp. 17\,452--17\,462.

\bibitem{chen2022learning}
W.-T. Chen, Z.-K. Huang, C.-C. Tsai \emph{et~al.}, ``Learning multiple adverse weather removal via two-stage knowledge learning and multi-contrastive regularization: Toward a unified model,'' in \emph{Proc. IEEE Conf. Comput. Vis. Pattern Recog.}, 2022, pp. 17\,653--17\,662.

\bibitem{lin2023lmqformer}
J.~Lin, N.~Jiang, Z.~Zhang \emph{et~al.}, ``Lmqformer: A laplace-prior-guided mask query transformer for lightweight snow removal,'' \emph{IEEE Trans. Circuits Syst. Video Technol.}, vol.~33, no.~11, pp. 6225--6235, 2023.

\bibitem{he2010DCP}
K.~He, J.~Sun, and X.~Tang, ``Single image haze removal using dark channel prior,'' \emph{IEEE Trans. Pattern Anal. Mach. Intell.}, vol.~33, no.~12, pp. 2341--2353, 2010.

\bibitem{li2017aod}
B.~Li, X.~Peng, Z.~Wang \emph{et~al.}, ``Aod-net: All-in-one dehazing network,'' in \emph{IEEE Conf. Comput. Vis. Pattern Recognit.}, 2017, pp. 4770--4778.

\bibitem{wu2021contrastive}
H.~Wu, Y.~Qu, S.~Lin \emph{et~al.}, ``Contrastive learning for compact single image dehazing,'' in \emph{Proc. IEEE Conf. Comput. Vis. Pattern Recog.}, 2021, pp. 10\,551--10\,560.

\bibitem{guo2020zero}
C.~Guo \emph{et~al.}, ``Zero-reference deep curve estimation for low-light image enhancement,'' in \emph{Proc. IEEE Conf. Comput. Vis. Pattern Recog.}, 2020, pp. 1780--1789.

\bibitem{cui2022you}
Z.~Cui \emph{et~al.}, ``You only need 90k parameters to adapt light: a light weight transformer for image enhancement and exposure correction,'' \emph{arXiv preprint arXiv:2205.14871}, 2022.

\bibitem{Mark2010VOC}
M.~Everingham, L.~Van~Gool, C.~K. Williams \emph{et~al.}, ``The pascal visual object classes (voc) challenge,'' \emph{Int. J. Comput. Vis.}, vol.~88, pp. 303--338, 2010.

\bibitem{quan2021removing}
R.~Quan, X.~Yu, Y.~Liang \emph{et~al.}, ``Removing raindrops and rain streaks in one go,'' in \emph{Proc. IEEE Conf. Comput. Vis. Pattern Recog.}, 2021, pp. 9147--9156.

\bibitem{chen2021CSD}
W.-T. Chen \emph{et~al.}, ``All snow removed: Single image desnowing algorithm using hierarchical dual-tree complex wavelet representation and contradict channel loss,'' in \emph{Proc. IEEE/CVF Int. Conf. Comput. Vis.}, 2021, pp. 4196--4205.

\bibitem{li2018RTTS}
B.~Li, W.~Ren, D.~Fu \emph{et~al.}, ``Benchmarking single-image dehazing and beyond,'' \emph{IEEE Trans. Image Process.}, vol.~28, no.~1, pp. 492--505, 2018.

\bibitem{Exdark}
Y.~P. Loh and C.~S. Chan, ``Getting to know low-light images with the exclusively dark dataset,'' \emph{Comput. Vis. Image Underst.}, vol. 178, pp. 30--42, 2019.

\end{thebibliography}

\end{document}